\documentclass[letterpaper, 10 pt, conference]{ieeeconf}  

\IEEEoverridecommandlockouts                              

\usepackage{amsmath} 
\usepackage{amssymb}  
\usepackage{multicol} 
\usepackage{multirow} 
\usepackage{makecell} 
\usepackage{booktabs} 
\usepackage{hyperref}
\usepackage{graphicx}
\usepackage{amssymb}
\usepackage{caption}
\usepackage{tabularx}
\usepackage{multirow}
\usepackage{array} 
\newcolumntype{Y}{>{\centering\arraybackslash}X}

\usepackage{subcaption} 
\usepackage{bm}
\hypersetup{
    colorlinks=true,
    urlcolor=blue
}

\RequirePackage{xspace}
\newcommand{\MODEL}{ABot-Explorer\xspace}

\title{\LARGE \bf
Explore Like Humans: Autonomous Exploration with Online SG-Memo Construction for Embodied Agents
}

\author{
Xu Chen$^{1*}$ \quad
Shichao Xie$^{1*}$ \quad
Zhining Gu$^{1}$ \quad
Lu Jia$^{1}$ \quad
Minghua Luo$^{1}$ \quad
Fei Liu$^{1}$ \quad
Zedong Chu$^{1\dagger}$ \\
Yanfen Shen$^{1}$ \quad
Xiaolong Wu$^{1}$ \quad
Mu Xu$^{1}$%
\thanks{$^{1}$Amap, Alibaba Group, China.}%
\thanks{$^{*}$Equal contribution.}%
\thanks{$^{\dagger}$Corresponding author:
\href{mailto:chuzedong.czd@alibaba-inc.com}{chuzedong.czd@alibaba-inc.com}.}%
}

\begin{document}

\maketitle
\thispagestyle{empty}
\pagestyle{empty}

\begin{abstract}

Constructing structured spatial memory is essential for enabling long-horizon reasoning in complex embodied navigation tasks. Current memory construction predominantly relies on a decoupled, two-stage paradigm: agents first aggregate environmental data through exploration, followed by the offline reconstruction of spatial memory. However, this post-hoc and geometry-centric approach precludes agents from leveraging high-level semantic intelligence, often causing them to overlook navigationally critical landmarks (e.g., doorways and staircases) that serve as fundamental semantic anchors in human cognitive maps. To bridge this gap, we propose \MODEL, a novel active exploration framework that unifies memory construction and exploration into an online, RGB-only process. At its core, \MODEL leverages Large Vision-Language Models (VLMs) to distill Semantic Navigational Affordances (SNA), which act as cognitive-aligned anchors to guide the agent’s movement. By dynamically integrating these SNAs into a hierarchical SG-Memo, \MODEL mirrors human-like exploratory logic by prioritizing structural transit nodes to facilitate efficient coverage. To support this framework, we contribute a large-scale dataset extending InteriorGS with SNA and SG-Memo annotations. Experimental results demonstrate that \MODEL significantly outperforms current state-of-the-art methods in both exploration efficiency and environment coverage, while the resulting SG-Memo is shown to effectively support diverse downstream tasks. Code is available at \url{https://github.com/amap-cvlab/ABot-Explorer}.

\end{abstract}


\section{INTRODUCTION}

In Embodied AI, constructing structured spatial memory is fundamental for mitigating long-horizon reasoning bottlenecks in complex indoor environments~\cite{zhang2026spatialnav, abot-n0, astranav-world, nav-r2, navforesee, socialnav, omninav, ce-nav}. A robust global memory provides the necessary grounding for diverse downstream tasks, such as Vision-and-Language Navigation (VLN)~\cite{zhang2026spatialnav} and Object Goal Navigation (ObjectNav)~\cite{zhou2025fsr}. While recent advancements leverage human-provided demonstrations or pre-collected trajectories to build such representations~\cite{zhang2026spatialnav,chen2025astra, chiang2024mobility}, achieving autonomous exploration that dynamically constructs a comprehensive global memory remains a critical and active research frontier.

Currently, autonomous memory construction predominantly relies on a decoupled, two-stage paradigm: agents first execute exploration trajectories to collect environmental data~\cite{bircher2016nbvp, cao2021tare}, followed by the offline generation of spatial memory. Within this pipeline, Scene Graphs (SGs) have emerged as a predominant paradigm for the abstraction of spatial memory information, as they encapsulate richer spatial-semantic information and hierarchical abstractions, providing a more robust foundation for grounding complex downstream reasoning~\cite{gu2024conceptgraphs}. However, this disjointed two-stage approach introduces two primary limitations. First, during exploration, agents typically depend on explicit geometric occupancy grids or topological graphs~\cite{li2025gvd, niu2025skeleton}, remain geometry-centric and lack semantic depth. Consequently, they frequently overlook navigationally critical nodes---such as staircases or narrow doorways---that are geometrically ambiguous but can be robustly identified through visual semantics. Second, although SGs are purposefully designed to mirror human cognitive abstractions, their post-hoc construction in current systems precludes the agent from leveraging this high-level semantic intelligence during the exploration phase, which missed an opportunity to improve exploration efficiency and coverage.

To overcome these limitations, we introduce \MODEL, a novel active exploration framework relying solely on RGB inputs. By leveraging Large Vision-Language Models (VLMs), \MODEL distills \textbf{Semantic Navigational Affordances} (SNA) to identify critical transit nodes---such as doorways, intersections, and staircases---that form the structural backbone of human-like exploration. These SNAs serve as the fundamental anchors for \textbf{SG-Memo} (\textbf{S}cene \textbf{G}raph \textbf{Memo}ry), a hierarchical memory representation designed to provide a structured, reasoning-ready foundation. By dynamically organizing SNA nodes into this representation concurrent with exploration, SG-Memo enables complex semantic and spatial reasoning within an online-constructed framework. Notably, \MODEL can be seamlessly interfaced with off-the-shelf embodied policy models~\cite{chu2026abot, liu2025navforesee} to execute low-level exploratory navigation primitives. Beyond exploration, SG-Memo provides a robust reasoning-ready knowledge that is directly applicable to diverse downstream embodied navigation tasks.

\begin{figure*}[t] 
    \centering
    \includegraphics[width=\textwidth]{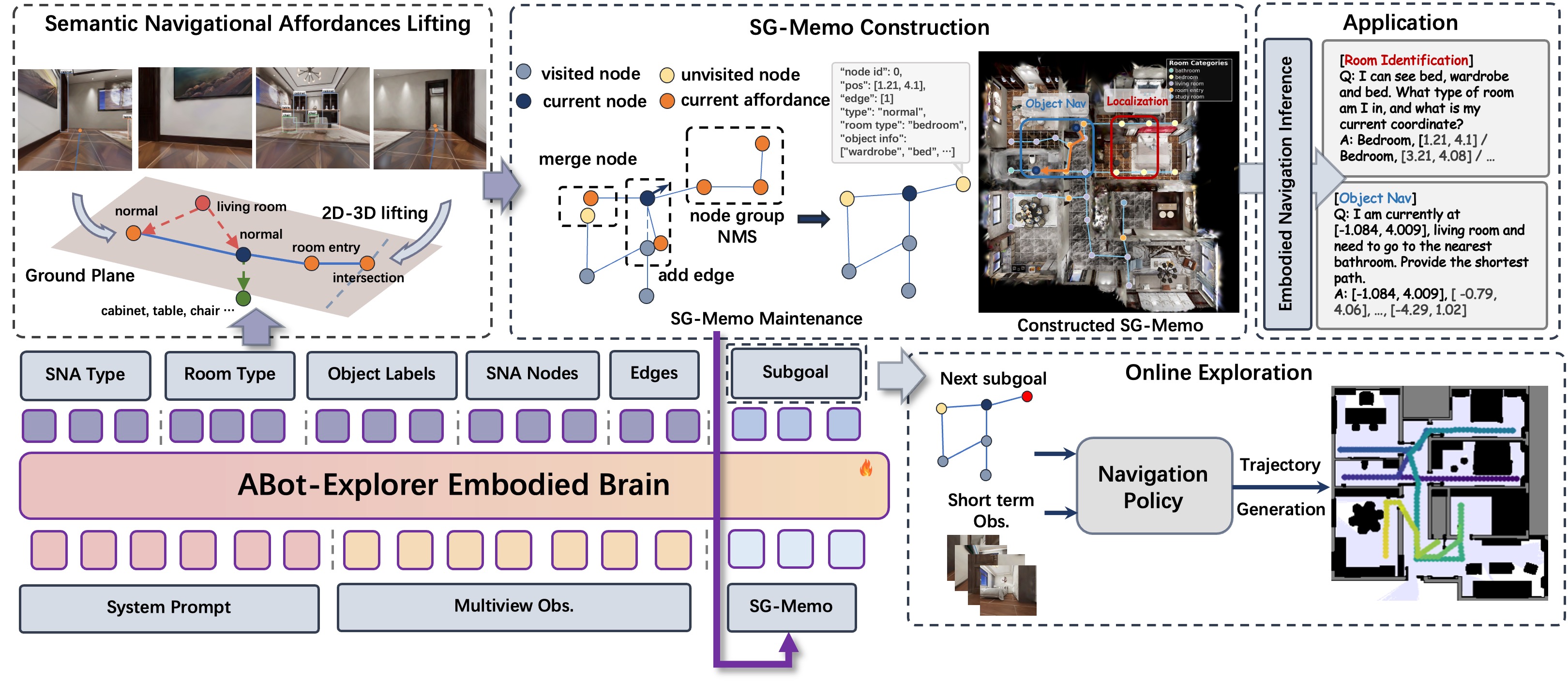} 
    \caption{Overview of our exploration and online SG-Memo construction pipeline: the \MODEL predicts SNA and scene semantics from RGB, updates a global SG-Memo, selects exploration subgoals, and executes them via a navigation module.}
    \vspace{-3mm}
    \label{fig:system_overview}
\end{figure*}

In summary, our main contributions are as follows:

\begin{itemize}
\item \textbf{Cognitive-Aligned Spatial Memory:} We introduce SG-Memo, a hierarchical memory representation built upon VLM-distilled SNA. By aligning agent perception with human cognitive abstractions, SG-Memo enables semantic-spatial reasoning and bridges the gap between raw perception and high-level navigation.

\item \textbf{RGB-Only Active Exploration:} We propose \MODEL, an RGB-only active exploration framework centered on visual semantics. By prioritizing navigationally critical bottlenecks over pure geometric connectivity, \MODEL achieves robust, human-like environmental coverage and superior exploration efficiency in complex scenarios.

\item \textbf{Comprehensive Dataset and Downstream Benchmark:} We curate a large-scale dataset extending InteriorGS, encompassing over 1,000 indoor scenes augmented with SNA and occlusion-aware visible object annotations. Furthermore, we establish a rigorous quantitative evaluation protocol to validate the practical utility of our dynamically constructed SG-Memo through downstream Embodied navigation tasks.
\end{itemize}

The model, datasets, and the proposed benchmark will be open-sourced to facilitate further research and iterative improvements in this domain.

\section{RELATED WORKS}

\subsection{Autonomous Exploration in Unknown Environments}

Traditional autonomous exploration relies heavily on explicit geometric occupancy maps, selecting actions via frontier reasoning or information-gain objectives~\cite{bircher2016nbvp, cao2021tare}. While learning-based policies~\cite{chen2025gleam} have been introduced to improve routing efficiency, dense occupancy-based methods often struggle with local optima in large-scale environments. Consequently, topological abstractions have emerged to facilitate global reasoning. For instance, GVD-TG~\cite{li2025gvd} utilizes Hierarchical Generalized Voronoi Diagrams to preserve narrow passages, while STGPlanner~\cite{niu2025skeleton} generates skeleton graphs directly from occupancy maps to maintain connectivity. More recently, predictive planners have attempted to anticipate unobserved structures; PIPE Planner~\cite{baek2025pipe} predicts occupancy grids, while CogniPlan~\cite{wang2025cogniplan} and P$^2$Explore~\cite{song2025p} infer room layouts to construct connectivity topologies.

Despite these advancements, existing pipelines remain fundamentally geometry-centric, struggling to perceive geometrically ambiguous but semantically critical bottlenecks like doorways and stairs. Even visual frontier-based approaches~\cite{alama2025rayfronts, hutter2025frontiernet} frequently waste exploration budgets on non-navigable boundaries (e.g., walls or furniture) because they lack semantic awareness. In contrast to these redundant geometric paradigms, \MODEL bypasses traditional frontiers entirely. By leveraging VLMs to distill SNA, we explicitly align exploration targets with the structural backbone of the scene. This semantic-centric strategy ensures the robust identification of transit nodes, significantly enhancing both exploration efficiency and environmental coverage.

\begin{figure*}[t] 
    \centering
    \includegraphics[width=\textwidth]{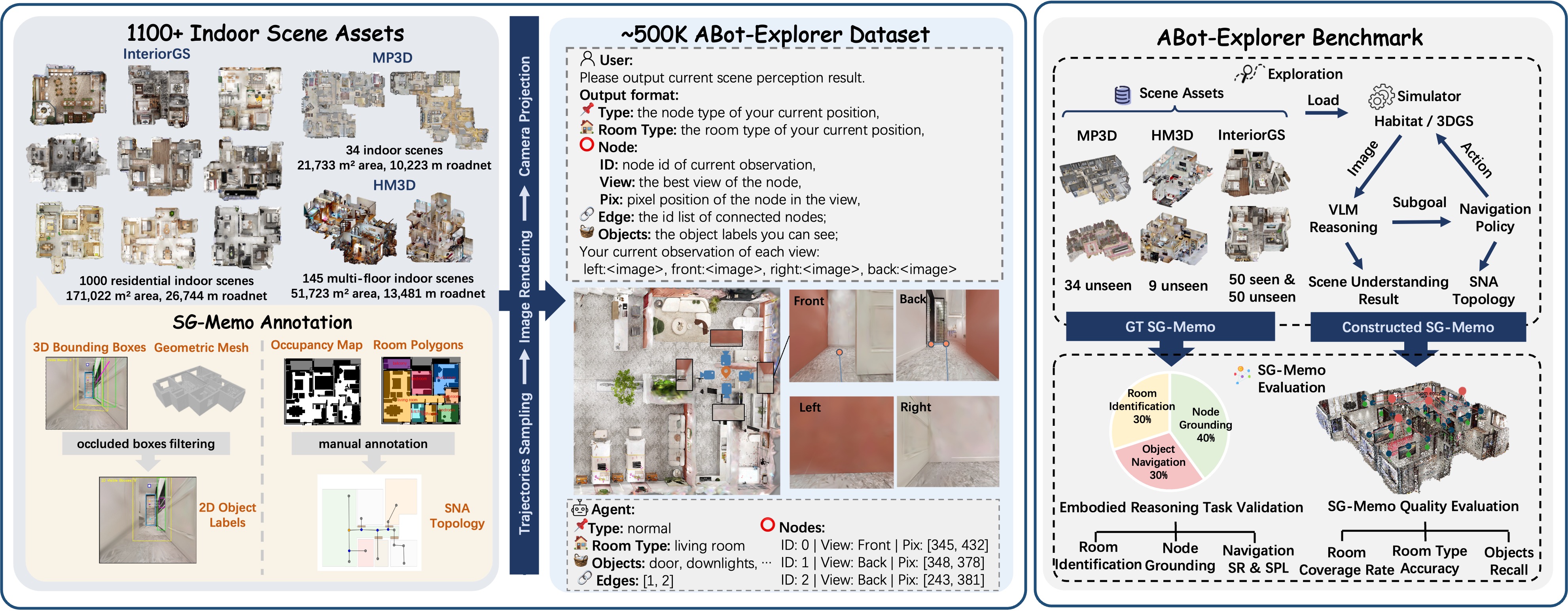} 
    \caption{Overview of the ABot-Explorer dataset and benchmark. We curate 1,179 indoor scenes across InteriorGS (1,000), HM3D~\cite{ramakrishnan2021hm3d} (145), and MP3D~\cite{chang2017matterport3d} (34), generating approximately 500K VQA pairs for training. The ABot-Explorer benchmark is established using 34 unseen MP3D, 9 unseen HM3D, and 100 InteriorGS scenes (comprising 50 seen and 50 unseen environments) to evaluate model performance and generalization.}
    \vspace{-3mm}
    \label{fig:dataset_benchmark}
\end{figure*}

\subsection{Scene Graph Construction for Embodied Navigation}

Graph-based representations provide explicit topology and structured memory crucial for long-horizon navigation~\cite{chiang2024mobility, chen2025astra, zhou2025fsr, zhang2026spatialnav}. Despite their utility, constructing these SGs predominantly follows a decoupled, offline paradigm. Building on early models~\cite{armeni20193d}, modern approaches leverage foundation models for open-vocabulary mapping, LLM grounding, and dynamic maintenance~\cite{werby2024hovsg, gu2024conceptgraphs, shafiullah2023clip, peng2023openscene, ju2024dovsg}. However, this disjointed paradigm fundamentally severs graph construction from active exploration. Consequently, the rich semantic context of the SG cannot be utilized to guide the agent, inherently bottlenecking exploration efficiency, while extracting the final structure still necessitates computationally heavy offline post-processing.

To address offline limitations, some methods attempt to incrementally construct graphs online. For instance, VGM~\cite{kwon2021visual}, TSGM~\cite{kim2023topological}, MemoNav~\cite{li2024memonav}, Smartway~\cite{shi2025smartway}, and ETPNav~\cite{an2024etpnav} dynamically build connected topological graphs to help agents reason about explored and unexplored regions. The underlying nodes for these graphs are typically provided by discretizing the environment~\cite{liu2025fine} or using dedicated visual waypoint predictors~\cite{hong2022bridging}. However, these waypoints are primarily trained on purely geometric features within traversable free space. More recently, MSGNav~\cite{huang2025msgnav} and SG-Nav~\cite{yin2024sg} have begun constructing scene graphs online to aid LLM reasoning in embodied navigation tasks. Yet, during active exploration, they still fundamentally rely on occupancy-based geometric frontiers as their navigational nodes, failing to leverage structured semantic nodes to guide the exploration process itself.

In contrast, our SG-Memo represents a paradigm shift toward cognitive-aligned spatial memory. Unlike geometry-centric waypoints, SG-Memo is built upon SNA nodes. By fine-tuning VLMs to distill these structured nodes online, \MODEL not only simplifies dynamic graph maintenance but also provides a reasoning-ready knowledge substrate that can be directly leveraged for diverse downstream embodied tasks.

\section{METHOD}

\subsection{Problem Statement}

The goal of this work is to enable a ground robot to autonomously
explore an unknown environment $\Omega \subset \mathbb{R}^3$ and dynamically construct a SG-Memo. We formulate the memory as a graph
$\mathcal{G}=(\mathcal{V},\mathcal{E})$, where the vertex set $\mathcal{V}$ consists entirely of Semantic Navigational Affordance (SNA) nodes. Rather than treating rooms and objects as separate nodes, each vertex $v \in \mathcal{V}$ encapsulates rich hierarchical semantic attributes---specifically, its SNA type, the associated room category, and visible object instances. Meanwhile, the edge set $\mathcal{E}$ encodes the navigable topological connectivity between these SNA nodes. Starting from an initial pose $\mathbf{x}_{init}$, the robot executes a pose sequence $\mathbf{X}=\{(\mathbf{p}_t,\mathbf{q}_t)\}_{t=1}^{T}$, with $\mathbf{p}_t\in\mathbb{R}^3$ and $\mathbf{q}_t\in\mathbb{SO}(3)$, to collect visual observations. In line with mainstream paradigms, the primary objective is to maximize the planar spatial coverage $|\Omega_{known}| \subseteq |\Omega|$ of the environment. However, as purely geometric coverage is insufficient for downstream reasoning, we introduce a new objective: maximizing the recall of SNA nodes. Given a ground-truth set of nodes $\mathcal{N}_{GT}=\{n_i\}_{i=1}^{M}$ at locations $\mathbf{s}(n_i)\in\Omega$, the agent must maximize the discovered subset $|\mathcal{N}_{known}| \subseteq |\mathcal{N}_{GT}|$.

\subsection{System Overview}

An overview of the proposed system is illustrated in Fig.~\ref{fig:system_overview}. The core component is our \MODEL, which operates exclusively on RGB streams to directly predict SNAs alongside hierarchical scene semantics (e.g., room categories and object instances). These 2D detections are subsequently projected into a Bird's-Eye-View (BEV) space to instantiate local structural nodes. The \MODEL dynamically maintains the SG-Memo of these SNA nodes and selects the optimal unvisited target. Once assigned, a navigation policy such as ABot-N0~\cite{chu2026abot} takes over for local obstacle avoidance and continuous path execution. By seamlessly coupling this high-level semantic reasoning with low-level policy control, the system natively aggregates reasoning results across traversed nodes to dynamically construct a comprehensive SG-Memo online.

\subsection{Semantic Navigational Affordance}

Distinct from conventional exploration algorithms that rely on geometric frontiers, we ground our exploration in SNA. These affordances are sparse, structurally meaningful landmarks, encompassing four types: room entries, stairs, intersections and normal nodes. Upon reaching each SNA, the agent leverages the model to predict and encapsulate the following hierarchical semantic attributes:

\begin{itemize}
    \item \textbf{SNA Type:} The model explicitly classifies the specific structural category of the current nodes.
    \item \textbf{Room Category:} Each node is assigned a semantic room label (e.g., bedroom, kitchen) to provide high-level spatial context and anchor the global layout.
    \item \textbf{Visible Objects:} Nodes are further enriched with nearby salient objects. We utilize a curated vocabulary of navigation-relevant object categories to ground the local visual semantics in SG-Memo.
\end{itemize}

We formalize the extraction of these SNA nodes and their interconnectivity as a direct reasoning task driven by a Vision-Language Model (VLM). At time step $t$, given an observation comprising $M$ multiview RGB images $\mathcal{I}_t = \{I_{t,k}\}_{k=1}^M$ where $I_{t,k} \in \mathbb{R}^{H \times W \times 3}$, alongside a task-specific language prompt $\mathcal{L}$, the VLM predicts a local 2D scene graph $\mathcal{G}^{2D}_t = (\mathcal{V}^{2D}_t, \mathcal{E}_t)$. The vertex set $\mathcal{V}^{2D}_t = \{v^{2D}_i\}_{i=1}^N$ represents the locally detected SNA nodes. Each node $v^{2D}_i$ is parameterized in pixel coordinates. The edge set $\mathcal{E}_t$ explicitly models local traversability; an edge $e_{ij} = (v_i, v_j) \in \mathcal{E}_t$ exists if and only if a navigable path connects them on the floor plane.

\subsection{Dataset Construction and Training Procedure}

To generate the required training data, we curated a comprehensive dataset based on the InteriorGS asset, encompassing 1,000 indoor scenes equipped with 3DGS reconstructions, occupancy maps, polygonal layouts, collision meshes~\cite{miao2025towards}, and object labels (illustrated in Fig.~\ref{fig:dataset_benchmark}). Building upon this foundation, we enriched the assets to align with our SNA formulation. Specifically, we manually annotated the SNA nodes and their navigable edges directly on the occupancy maps, and systematically assigned room categories via point-in-polygon queries against the layout annotations. We then sampled  poses on SNA topology within each scene and rendered egocentric images along the paths, projecting the above annotations into the corresponding views to construct the training dataset.

To curate object-level supervision, we observed that directly projecting 2D map annotations into the camera view introduces severe occlusion artifacts. To guarantee visible ground truth for the VLM, we perform an occlusion-culling pipeline: we project coarse object candidates into the image plane and then run ray-casting checks against 3D collision meshes to filter out occluded instances. Finally, visible object labels undergo manual verification, ensuring high-fidelity multimodal supervision.

We fine-tune Qwen2.5-VL-3B via Supervised Fine-Tuning (SFT) to explicitly ground SNA into the model's spatial reasoning. As illustrated in Fig.~\ref{fig:dataset_benchmark}, rather than treating reasoning as isolated tasks, the model is trained to jointly localize SNA nodes at their original image resolution, infer traversable connectivity, and identify hierarchical semantics. Formally, conditioned on the current multiview observation $\mathcal{I}_t$ and the task-specific language prompt $\mathcal{L}$, the VLM generates a serialized sequence $y = \{y_1, \dots, y_T\}$ that decodes into the local 2D graph $\mathcal{G}^{2D}_t$. We optimize the network weights $\theta$ using the standard autoregressive negative log-likelihood loss:
\begin{equation}
\mathcal{L}_{\mathrm{SFT}}(\theta) = -\sum_{j=1}^{T} \log p_\theta\!\left(y_j \mid \mathcal{I}_t, \mathcal{L}, y_{<j}\right),
\end{equation}
where $p_\theta(y_j \mid \mathcal{I}_t, \mathcal{L}, y_{<j})$ denotes the probability of predicting the $j$-th token given the multimodal inputs and all preceding tokens $y_{<j}$.

\subsection{Autonomous Exploration via Online SG-Memo Construction}
\label{sec:Exploration with Topological}

During active exploration, the agent dynamically maintains a global SG $\mathcal{G}_t=(\mathcal{V}_t,\mathcal{E}_t)$, initialized with the robot's starting pose $\mathbf{v}_0$. To integrate $\mathcal{G}^{2D}_t$ into the global SG-Memo, we map the 2D vertices $\mathcal{V}^{2D}_t$ to the BEV frame. We adopt a geometric approach based on Inverse Perspective Mapping (IPM). Assuming the robot navigates on a planar surface, we directly project the 2D pixel coordinates onto the floor plane by leveraging the camera intrinsic matrix and the agent's calibrated extrinsic pose. Empirically, we find that this straightforward IPM projection yields sufficiently accurate 3D node localization. Through this streamlined geometric projection, the visually predicted local graph is lifted into a local 3D graph $\mathcal{G}^{l}_t$.

Newly detected SNA nodes $u \in \mathcal{V}^l_t$ are reconciled with the existing graph via a spatial threshold $\epsilon$. Matches with visited nodes trigger edge creation to the current robot node $v_c$, while matches with unvisited nodes are merged through weighted position averaging. To maintain graph sparsity, we prune candidates that lack semantic features or lie proximal to historical trajectories. Finally, a connectivity-aware Non-Maximum Suppression (NMS) resolves local redundancies by grouping interconnected nodes and retaining only the candidate closest to $v_c$, yielding an unvisited SNA set as exploration subgoal candidates.

Guided by the serialized SG-Memo in JSON format, the VLM serves as a high-level planner to select an exploration subgoal via a greedy-reasoning system prompt. This prompt encourages the agent to first prioritize unvisited SNA nodes aligned with its current heading to ensure local smoothness, then fall back to the most distant candidates to maximize global spatial coverage. Upon reaching a node, its hierarchical semantics are updated, and spatially redundant nodes are merged through radius-based clustering. To preserve the structural backbone of the environment, we retain representative nodes according to an SNA priority hierarchy (i.e., stairs $>$ room entries $>$ intersections $>$ normal nodes) and reconstruct their navigable connectivity. Finally, scene attributes are refined by aggregating cluster data---using majority voting for room categories and set unions for object entities.



\begin{figure*}[t]
    \centering
    \includegraphics[width=\textwidth]{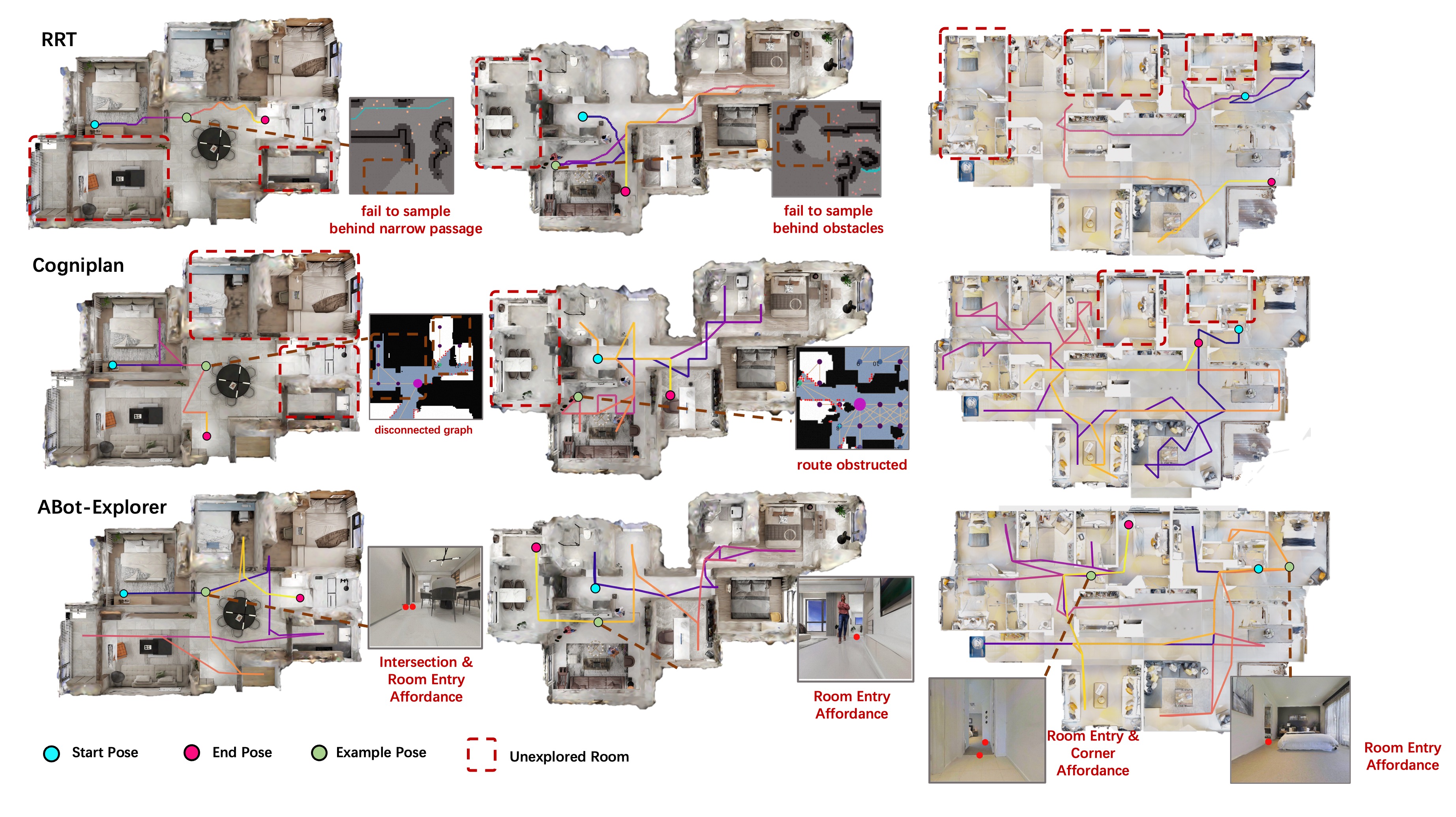}
    \caption{Visual comparison of exploration trajectories across different scenarios and algorithms, showing complete exploration paths in unseen InteriorGS scenes and a MP3D scene. Depth-dependent baselines such as RRT and Cogniplan fail to cover small rooms or narrow corridors.}
    \vspace{-3mm}
    \label{fig:explore_traj}
\end{figure*}

\section{Experiments}

\subsection{Experiments Setup}


To evaluate generalization, we benchmark on 143 indoor scenes. For InteriorGS, we report results on 50 seen (training) scenes and 50 held-out (unseen) scenes. We sample 10 random starting poses per InteriorGS scene, and use one start for each MP3D/HM3D scene. We assess exploration performance on InteriorGS, HM3D, and MP3D. In addition, we evaluate SG-Memo quality only on InteriorGS, where we further build an SG-Memo reasoning tasks and SG-Memo quality evaluation as Shown in Fig.~\ref{fig:dataset_benchmark}. Unless otherwise specified, our model is trained exclusively on the InteriorGS training set. Following the training setup, the agent uses four ground-parallel RGB cameras with $90^\circ$ yaw offsets for panoramic perception; each captures $720{\times}640$ images with a $110^\circ{\times}104^\circ$ FOV.

\subsection{Evaluation Metrics}

We comprehensively evaluate our system across two primary dimensions: the fundamental performance of exploration, and the structural quality and practical utility of the dynamically constructed .

\begin{table*}[t]
\centering
\caption{Comparison of exploration performance across scenarios and algorithms. Our method achieves the best node coverage and efficiency across all scenes, and the highest occupancy coverage and efficiency in most.}
\label{tab:main_results}

\small 
\scriptsize 
\setlength{\tabcolsep}{3.5pt} 
\renewcommand{\arraystretch}{0.95} 

\begin{tabular}{l l c c c c c}
\toprule
\multirow{2}{*}{\textbf{Scenario}} & \multirow{2}{*}{\textbf{Algorithm}} & \multirow{2}{*}{\textbf{use depth input}} &
\multicolumn{2}{c}{\textbf{Completeness}} & \multicolumn{2}{c}{\textbf{Efficiency}} \\
\cmidrule(lr){4-5} \cmidrule(lr){6-7}
& & & $CR_{\text{occ}}\uparrow$ & $CR_{\text{topo}}\uparrow$ & $AUC_{\text{occ}}\uparrow$ & $AUC_{\text{topo}}\uparrow$ \\
\midrule

\multirow{4}{*}{InteriorGS (Seen)} 
& RRT                       & $\times$      & 61\% & 72\% & 47\% & 56\% \\
& CogniPlan                 & $\times$      & 73\% & 77\% & 60\% & 60\% \\
& ABot-Explorer (nodes only)  & $\checkmark$  & 70\% & 85\% & 61\% & 70\% \\
& ABot-Explorer (full)        & $\checkmark$  & \textbf{86\%} & \textbf{96\%} & \textbf{84\%} & \textbf{79\%} \\
\midrule

\multirow{4}{*}{InteriorGS (Unseen)} 
& RRT                       & $\times$      & 57\% & 67\% & 43\% & 54\% \\
& CogniPlan                 & $\times$      & \textbf{81\%} & 86\% & \textbf{66\%} & 64\% \\
& ABot-Explorer (nodes only)  & $\checkmark$  & 71\% & 78\% & 53\% & 62\% \\
& ABot-Explorer (full)        & $\checkmark$  & 79\% & \textbf{89\%} & 60\% & \textbf{70\%} \\
\midrule

\multirow{4}{*}{HM3D (Unseen)} 
& RRT                       & $\times$      & 60\% & 64\% & 46\% & 44\% \\
& CogniPlan                 & $\times$      & 62\% & 65\% & 48\% & 45\% \\
& ABot-Explorer (nodes only)  & $\checkmark$  & 58\% & 61\% & 47\% & 40\% \\
& ABot-Explorer (full)        & $\checkmark$  & \textbf{71\%} & \textbf{71\%} & \textbf{55\%} & \textbf{54\%} \\
\midrule

\multirow{5}{*}{MP3D (Unseen)} 
& RRT                       & $\times$      & 21\% & 24\% & 12\% & 14\% \\
& GLEAM                     & $\times$      & 42\% & 48\% & 37\% & 41\% \\
& CogniPlan                 & $\times$      & 70\% & 69\% & 56\% & 49\% \\
& ABot-Explorer (nodes only)  & $\checkmark$  & 60\% & 58\% & 45\% & 44\% \\
& ABot-Explorer (full)        & $\checkmark$  & \textbf{86\%} & \textbf{85\%} & \textbf{65\%} & \textbf{62\%} \\
\bottomrule
\end{tabular}
\vspace{-3mm}
\end{table*}

\noindent\textbf{1. Exploration Performance} \\
To assess the effectiveness of the active exploration policy, we rely on two standard metrics:
\begin{itemize}
    \item \textbf{Completeness:} Measured via the geometric occupancy coverage rate ($\text{CR}_{\text{occ}}$) and the SNA node recall ($\text{CR}_{\text{topo}}$). A ground-truth node is deemed successfully discovered if the agent's trajectory intersects a $2\text{m}$ radius around its designated location.
    \item \textbf{Efficiency:} We measure exploration efficiency using the area under the coverage--path-length curve (AUC). For each modality, we integrate the coverage rate over the traveled distance and normalize by the final path length. 
\end{itemize}

\noindent\textbf{2. SG-Memo Quality and Downstream Utility} \\
To validate that our purely geometry-free exploration yields semantically rich representations, we benchmark the online-constructed SG-Memo against ground-truth annotations on unseen InteriorGS environments.
\begin{itemize}
    \item \textbf{Graph Construction Quality:} We assess SG-Memo fidelity by measuring room coverage, room-type accuracy, and node-level visible object recall.
    \item \textbf{Downstream Utility:} Provided with the serialized SG-Memo, an LLM performs spatial reasoning tasks across three dimensions: 
    (1) \textit{Room Identification}: accuracy of grounding linguistic scene descriptions to the correct room nodes; 
    (2) \textit{Node Grounding}: success rate of localizing prompt-described nodes within the correct room and a $2\,\mathrm{m}$ Euclidean threshold; 
    (3) \textit{Object Navigation}: SR and SPL for reaching target objects via SG-Memo-guided global path planning.
\end{itemize}

\subsection{Comparison of Exploration Performance with Existing Approaches}

Table~\ref{tab:main_results} presents a quantitative comparison of our approach against representative depth-based exploration baselines across various scenarios. These include the classical sampling-based planner, RRT, and two recent learning-based methods, GLEAM~\cite{chen2025gleam} and CogniPlan~\cite{wang2025cogniplan}. 
Quantitative results show that ABot-Explorer consistently outperforms all baselines on node-level completeness and efficiency ($CR_{\text{topo}}$, $AUC_{\text{topo}}$) across all scenarios, while achieving the best occupancy-based metrics ($CR_{\text{occ}}$, $AUC_{\text{occ}}$) in most cases. In the unseen InteriorGS setting, our occupancy scores are slightly below CogniPlan, reflecting an intentional trade-off: we prioritize SNA coverage over pure geometric expansion to better support scene-graph construction. Fig.~\ref{fig:explore_traj} further illustrates these differences. RRT-style methods suffer from inefficient random sampling, often getting stuck in narrow passages and terminating early in large MP3D scenes. CogniPlan's rule-based uniform node generation is brittle in complex layouts, frequently yielding disconnected subgraphs and blocked routes. In contrast, ABot-Explorer uses affordance detection with edge prediction under partial views to plan semantically grounded, well-connected paths and achieve more complete exploration.

\begin{figure}[h]
    \centering
    \includegraphics[width=8cm, keepaspectratio]{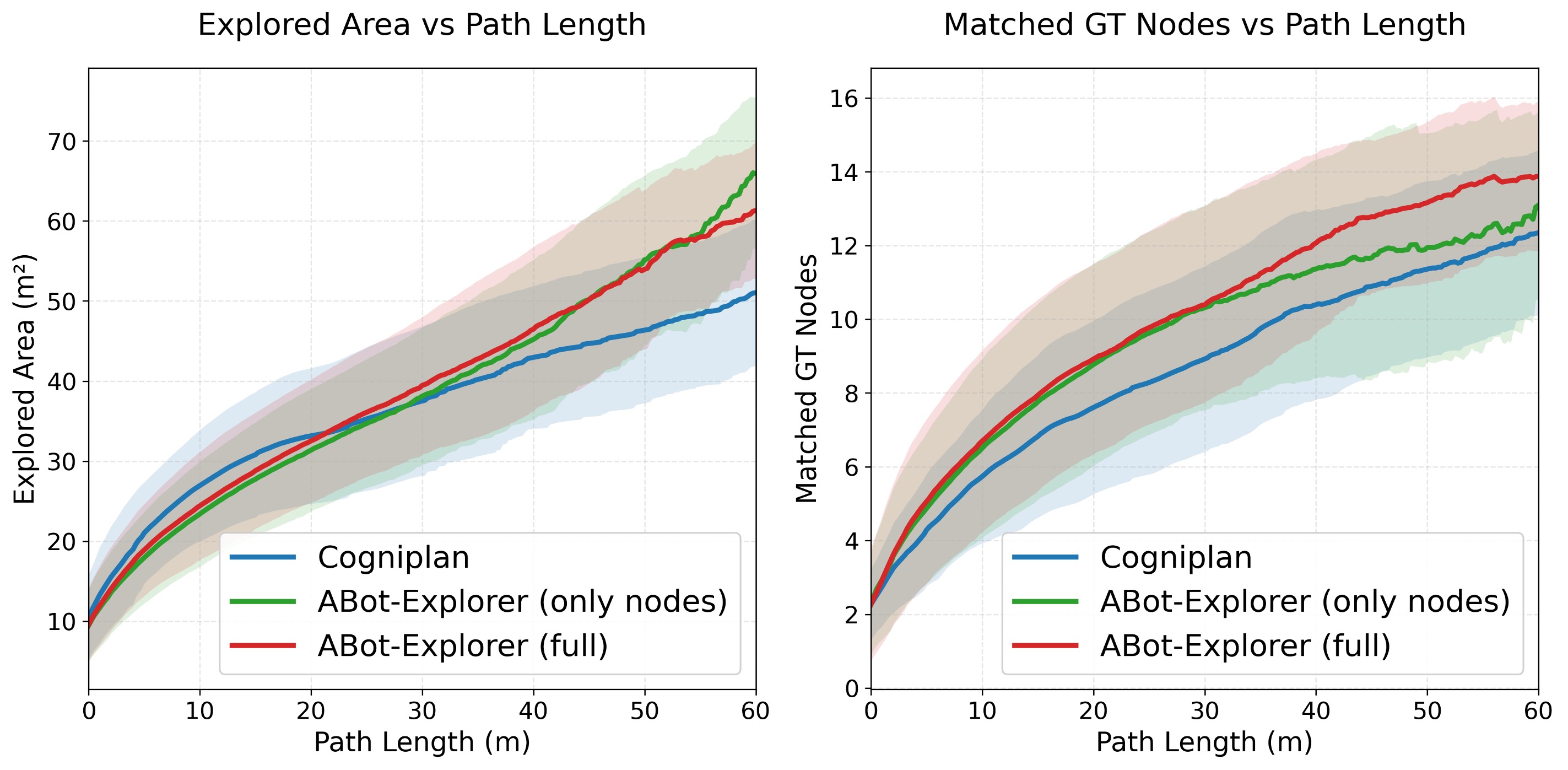}
    \caption{Occupancy and node coverage vs. trajectory length (0–60 m).}
    \vspace{-3mm}
    \label{fig:explore_curve}
\end{figure}

To further compare exploration efficiency, we report detailed coverage-versus-trajectory-length growth curves in Fig.~\ref{fig:explore_curve}. Our method demonstrates superior exploration efficiency compared to Cogniplan. 

\begin{figure}[h]
    \centering
    \includegraphics[width=8cm, keepaspectratio]{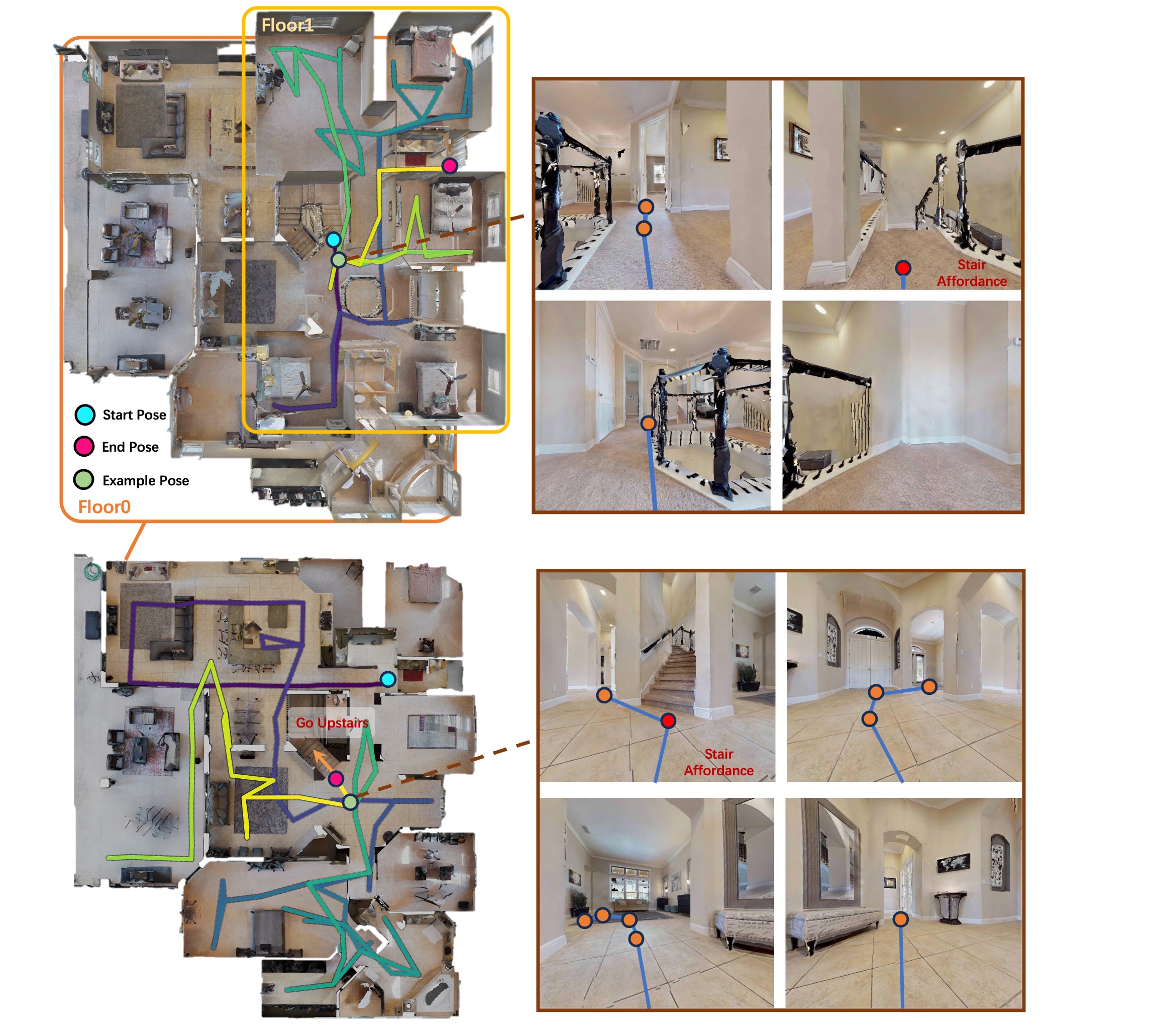}
    \caption{Cross-floor exploration results. We initialize the agent on the first floor of an HM3D scene.}
    \vspace{-3mm}
    \label{fig:multifloor_vis}
\end{figure}

Finally, we evaluate cross-floor exploration on HM3D. While geometric baselines like CogniPlan fail to distinguish stairs from occupancy grids, our model perceives stairs as SNAs by leveraging annotations from the HM3D multi-floor training set (136 scenes, with 9 held-out for evaluation). During inference, once a floor is explored, the agent targets a detected stair node and executes inter-floor transitions via navigation-policy instruction following. As shown in Fig.~\ref{fig:multifloor_vis} and Table~\ref{tab:comparison_multi_floor}, this semantic awareness enables robust exploration in complex multi-floor environments.

\subsection{SG-Memo Construction Results and Analysis}





Table~\ref{tab:sg_and_downstream} summarizes the results across three configurations:
(1) CogniPlan, which generates topologies through trajectory sparsification and DBSCAN post-processing, followed by Gemini-3-Pro for node-level object and room-type extraction;
(2) ABot-Explorer (traj), which leverages the exploration trajectory of ABot-Explorer but applies the identical post-processing pipeline as in (1) for semantic inference;
(3) Full ABot-Explorer, our unified framework featuring integrated online SG-Memo construction.



\begin{table}[t]
\centering
\caption{SG-Memo evaluation on unseen InteriorGS.}
\label{tab:sg_and_downstream}
\footnotesize

\small 
\scriptsize 
\setlength{\tabcolsep}{3.5pt} 
\renewcommand{\arraystretch}{0.95} 

\begin{tabularx}{\linewidth}{l *{3}{>{\centering\arraybackslash}X}}
\toprule
\multicolumn{4}{l}{\textbf{(a) \textsc{Graph Construction Quality}}} \\ 
\midrule
\textbf{Graph Source} & \textbf{Room Coverage} & \textbf{Room Type} & \textbf{Object Recall} \\
\midrule
CogniPlan            & 83\%  & 88\%  & 75\%  \\
ABot-Explorer (traj) & 90\%  & 82\%  & 80\%  \\
ABot-Explorer (full) & \textbf{92\%} & \textbf{90\%} & \textbf{81\%} \\
\bottomrule
\end{tabularx}

\vspace{3mm} 

\begin{tabularx}{\linewidth}{l *{4}{>{\centering\arraybackslash}X}}
\toprule
\multicolumn{5}{l}{\textbf{(b) \textsc{Downstream Embodied Reasoning Task}}} \\
\midrule
\textbf{Graph Source} & \textbf{Room \newline Identification} & \textbf{ObjectNav SR} & \textbf{ObjectNav SPL} & \textbf{Node Grounding} \\
\midrule
Ground-truth         & 96\% & 92\% & 92\% & 99\% \\
CogniPlan            & 64\% & 33\% & 31\% & 68\% \\
ABot-Explorer (traj) & 64\% & 46\% & 40\% & 60\% \\
ABot-Explorer (full) & \textbf{70\%} & \textbf{65\%} & \textbf{62\%} & \textbf{80\%} \\
\bottomrule
\end{tabularx}
\vspace{-1mm}
\end{table}

\begin{table}[t]
\centering
\caption{Ablation Study On Training Data.}
\label{tab:ablation_all}
\footnotesize 
\small 
\scriptsize 
\setlength{\tabcolsep}{3.5pt} 
\renewcommand{\arraystretch}{0.95} 

\begin{tabularx}{\linewidth}{l *{4}{>{\centering\arraybackslash}X}}
\toprule
\multirow{2}{*}{\textbf{Setting}} & \multicolumn{2}{c}{\textbf{Completeness}} & \multicolumn{2}{c}{\textbf{Efficiency}} \\
\cmidrule(lr){2-3} \cmidrule(lr){4-5}
& $CR_o\uparrow$ & $CR_t\uparrow$ & $AUC_o\uparrow$ & $AUC_t\uparrow$ \\
\midrule
Cogniplan           & 62\% & 65\% & 48\% & 45\% \\
Only HM3D           & 72\% & 74\% & 56\% & 60\% \\
Only InteriorGS     & 71\% & 71\% & 55\% & 54\% \\
Full                & \textbf{79\%} & \textbf{82\%} & \textbf{57\%} & \textbf{62\%} \\
\bottomrule
\end{tabularx}
\vspace{-3mm}
\end{table}

\begin{table}[t]
\centering
\caption{Multi-floor exploration evaluation on HM3D.}
\label{tab:comparison_multi_floor}
\footnotesize 
\setlength{\tabcolsep}{2pt}
\renewcommand{\arraystretch}{1.1}
\begin{tabularx}{\linewidth}{l *{4}{>{\centering\arraybackslash}X}}
\toprule
\multirow{2}{*}{\textbf{Setting}} & \multicolumn{2}{c}{\textbf{Completeness}} & \multicolumn{2}{c}{\textbf{Efficiency}} \\
\cmidrule(lr){2-3} \cmidrule(lr){4-5}
& $CR_o\uparrow$ & $CR_t\uparrow$ & $AUC_o\uparrow$ & $AUC_t\uparrow$ \\
\midrule
Cogniplan           & 35\% & 33\% & 27\% & 23\% \\
ours(single floor)      & 45\% & 42\% & 32\% & 32\% \\
ours       & \textbf{72\%} & \textbf{71\%} & \textbf{54\%} & \textbf{55\%} \\
\bottomrule
\end{tabularx}
\end{table}

\begin{figure}[t]
    \centering
    \includegraphics[width=8cm, keepaspectratio]{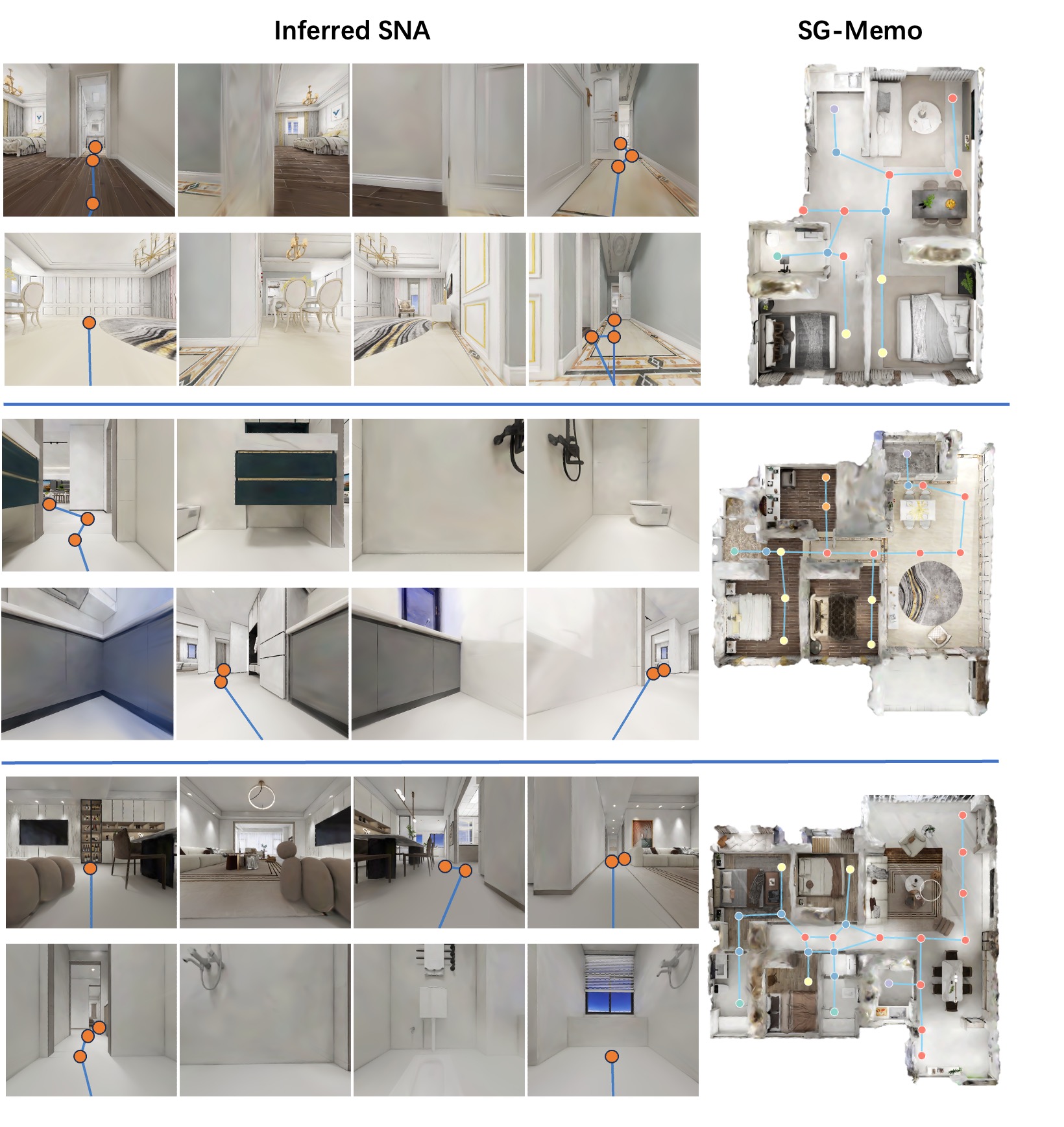}
    \caption{SG-Memo construction results in the simulated environment. We visualize the constructed SG-Memo and report the model's reasoning results on SNA.}
    \vspace{-2mm}
    \label{fig:sg_sim_vis}
\end{figure}

\begin{figure}[t]
    \centering
    \includegraphics[width=8cm, keepaspectratio]{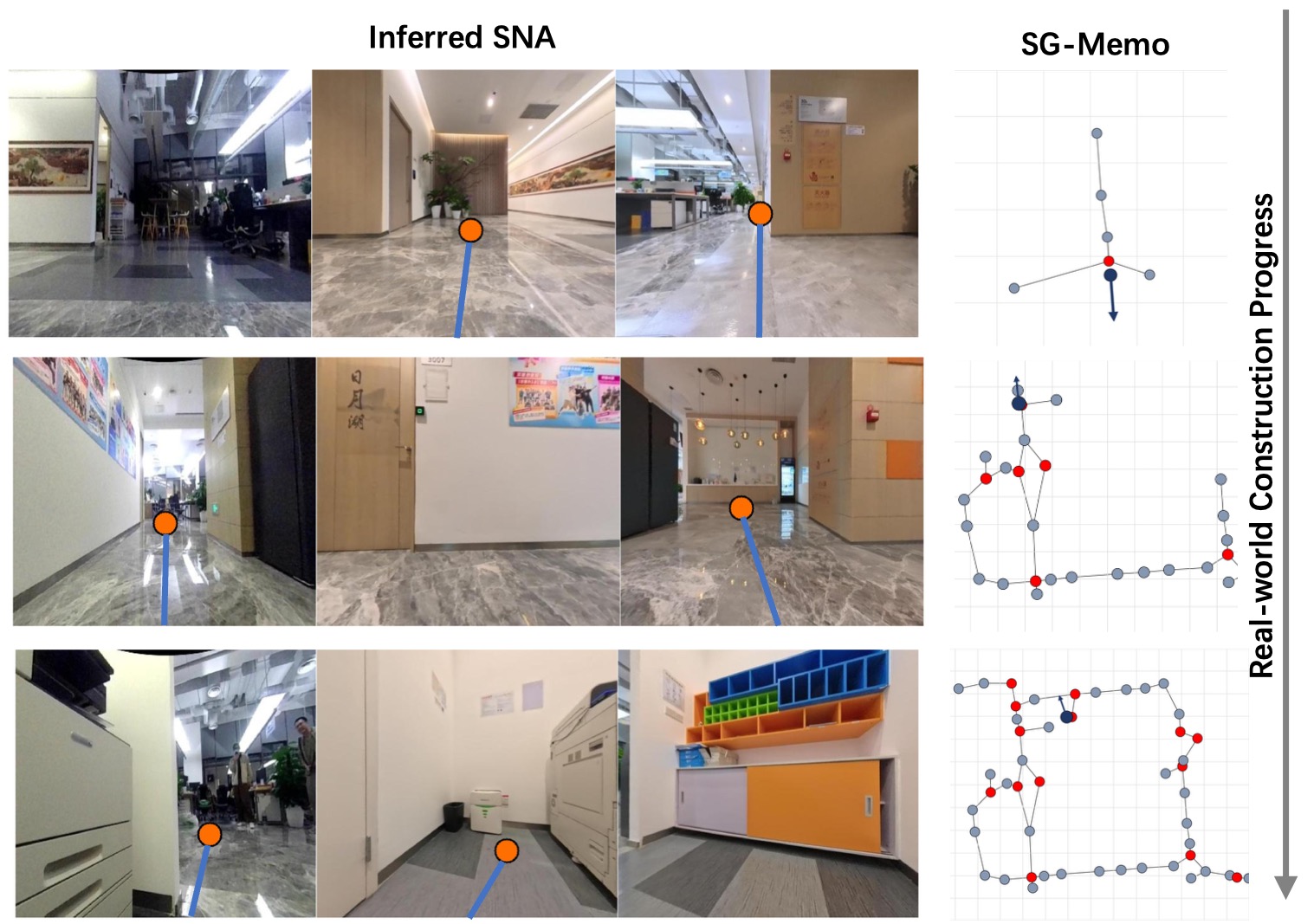}
    \caption{Online SG-Memo construction in real-world. }
    \vspace{-3mm}
    \label{fig:sg_real_vis}
\end{figure}


Compared with the two-stage methods, the full ABot-Explorer produces higher-quality SG-Memo both in graph construction quality (Table~\ref{tab:sg_and_downstream}(a)) and downstream task (Table~\ref{tab:sg_and_downstream}(b)). We attribute this to the fact that geometry-centric exploration baselines primarily optimize occupancy coverage and do not explicitly encourage node coverage; consequently, their trajectories often avoid entering rooms or reaching the ends of corridors, which reduces the global recall of both objects and room nodes. Higher room and object recall makes SG-Memo more complete, leading to higher success rates in downstream tasks. Moreover, our accurate node connectivity improves topological reasoning, significantly boosting ObjectNav performance.

\subsection{Ablation Studies}
We ablate training data and prediction targets to justify our design choices, referencing the 'nodes-only' variant (Table~\ref{tab:main_results}) and detailed results in Table~\ref{tab:ablation_all}. Specifically, the 'nodes-only' variant omits edge prediction between SNAs, instead assuming direct edges from the current node to all identified SNAs. To evaluate generalization and data quality, we compare models trained on our primary source, InteriorGS, against the HM3D training set.

Incorporating edge prediction substantially boosts performance across all scenarios. It enhances the VLM's spatial understanding and streamlines SG-Memo management through sparse connectivity, which significantly reduces the complexity of exploration decision-making. Regarding training data, the InteriorGS-only achieves performance comparable to HM3D-only, demonstrating its exceptional diversity and generalization despite the domain gap. We attribute the slight gap to the rendering-induced distribution shift between 3DGS and Habitat. Consequently, joint training yields the best results, where InteriorGS contributes broad generalization while HM3D provides critical domain alignment for the evaluation environment.



\subsection{Real-world Deployment and Qualitative Results}

We evaluate the robustness and generalization of our system through real-world deployment and extensive qualitative analysis in simulation.

\noindent\textbf{(1) Real-world Deployment.}
To validate the system's cross-domain applicability, we deployed ABot-Explorer on a MagicDog quadruped robot. The robot was equipped with a four-view camera rig, maintaining a sensor configuration identical to our training setup. The camera's extrinsic parameter with ground was calibrated. Upon reaching each exploration subgoal, the agent triggers a VLM inference by streaming the current multi-view observations via WebSocket to an offboard NVIDIA RTX 4090 GPU. As illustrated in Fig.~\ref{fig:sg_real_vis}, our framework successfully explore and construct SG-Memo in real-world indoor environments.

\noindent\textbf{(2) Qualitative Results in Simulation.}
We further provide visualizations of the online mapping process across several unseen InteriorGS scenes in Fig.~\ref{fig:sg_sim_vis}. Benefiting from SNA distilling and real-time SG-Memo construction, ABot-Explorer achieves human-like exploration strategy in complex, multi-room layouts.




\section{CONCLUSIONS}

This paper presents \MODEL, an online active exploration framework that unifies memory construction and navigation through SNA. Beyond its immediate performance, SG-Memo provides a reasoning-ready memory design, offering potential to serve as a foundational carrier for diverse downstream embodied tasks. By releasing our SNA-augmented dataset, we provide a foundation for advancing semantically-grounded scene understanding, enabling navigation agents to master complex, large-scale environments with human-aligned logic.




\bibliographystyle{IEEEtran}
\bibliography{IEEEabrv,reference}

@article{hutter2025frontiernet,
  title={FrontierNet: Learning Visual Cues to Explore},
  author={Hutter, Matthias and others},
  journal={arXiv preprint arXiv:2501.04597},
  year={2025}
}

@inproceedings{bircher2016nbvp,
  title={Receding horizon "next-best-view" planner for 3D exploration},
  author={Bircher, Andreas and Kamel, Mina and Alexis, Kostas and Oleynikova, Helen and Siegwart, Roland},
  booktitle={2016 IEEE international conference on robotics and automation (ICRA)},
  pages={1462--1468},
  year={2016},
  organization={IEEE}
}

@article{an2024etpnav,
  title={Etpnav: Evolving topological planning for vision-language navigation in continuous environments},
  author={An, Dong and Wang, Hanqing and Wang, Wenguan and Wang, Zun and Huang, Yan and He, Keji and Wang, Liang},
  journal={IEEE Transactions on Pattern Analysis and Machine Intelligence},
  year={2024},
  publisher={IEEE}
}

@article{zhang2026spatialnav,
  title={SpatialNav: Leveraging Spatial Scene Graphs for Zero-Shot Vision-and-Language Navigation},
  author={Zhang, Jiwen and Li, Zejun and Wang, Siyuan and Shi, Xiangyu and Wei, Zhongyu and Wu, Qi},
  journal={arXiv preprint arXiv:2601.06806},
  year={2026}
}

@inproceedings{kim2023topological,
  title={Topological Semantic Graph Memory for Image-Goal Navigation},
  author={Kim, Nuri and Kwon, Obin and Oh, Songhwai},
  booktitle={Conference on Robot Learning (CoRL)},
  year={2023}
}

@inproceedings{werby2024hovsg,
  title={Hierarchical Open-Vocabulary 3D Scene Graphs for Language-Grounded Robot Navigation},
  author={Werby, Abdelrhman and Huang, Chenguang and Fadini, Martin and Burgard, Wolfram and Valada, Abhinav},
  booktitle={Robotics: Science and Systems (RSS)},
  year={2024}
}

@article{ju2024dovsg,
  title={Dynamic Open-Vocabulary 3D Scene Graphs for Long-term Language-Guided Mobile Manipulation},
  author={Ju, Zhijie and Zhang, Zhangjie and Deng, Jindong and Xiong, Yaoxian and Zhang, Jing and Xu, Yue and Wang, Qi and Yu, Dacheng},
  journal={IEEE Robotics and Automation Letters},
  year={2025}
}

@inproceedings{kwon2021visual,
  title={Visual graph memory with unsupervised representation for visual navigation},
  author={Kwon, Obin and Kim, Nuri and Choi, Yunho and Yoo, Hwiyeon and Park, Jeongho and Oh, Songhwai},
  booktitle={Proceedings of the IEEE/CVF international conference on computer vision},
  pages={15890--15899},
  year={2021}
}

@inproceedings{li2024memonav,
  title={Memonav: Working memory model for visual navigation},
  author={Li, Hongxin and Wang, Zeyu and Yang, Xu and Yang, Yuran and Mei, Shuqi and Zhang, Zhaoxiang},
  booktitle={Proceedings of the IEEE/CVF Conference on Computer Vision and Pattern Recognition},
  pages={17913--17922},
  year={2024}
}

@article{chen2025gleam,
  author  = {Chen, Xiao and Wang, Tai and Li, Quanyi and Huang, Tao and Pang, Jiangmiao and Xue, Tianfan},
  title   = {{GLEAM}: Learning Generalizable Exploration Policy for Active Mapping in Complex {3D} Indoor Scenes},
  journal = {arXiv preprint arXiv:2505.20294},
  year    = {2025}
}

@inproceedings{armeni20193d,
  title={3D Scene Graph: A Structure for Unified Semantics, 3D Space, and Camera},
  author={Armeni, Iro and He, Zhi-Yang and Gwak, JunYoung and Zamir, Amir R and Fischer, Martin and Malik, Jitendra and Savarese, Silvio},
  booktitle={Proceedings of the IEEE/CVF International Conference on Computer Vision (ICCV)},
  pages={5664--5673},
  year={2019}
}

@inproceedings{shafiullah2023clip,
  title={CLIP-Fields: Weakly Supervised Semantic Fields for Robotic Memory},
  author={Shafiullah, Nur Muhammad Mahi and Cui, Zichen and Altanzaya, Ariuntuya and Pinto, Lerrel},
  booktitle={Robotics: Science and Systems (RSS)},
  year={2023}
}

@inproceedings{peng2023openscene,
  title={OpenScene: 3D Scene Understanding with Open Vocabularies},
  author={Peng, Songyou and Genova, Kyle and Jiang, Chane and Tagliasacchi, Andrea and Hormung, Marc and Funkhouser, Thomas and Tang, Siyu},
  booktitle={Proceedings of the IEEE/CVF Conference on Computer Vision and Pattern Recognition (CVPR)},
  year={2023}
}

@article{ramakrishnan2021hm3d,
  title={Habitat-Matterport 3D Dataset (HM3D): 1000 Large-scale 3D Environments for Embodied AI},
  author={Ramakrishnan, Santhosh K and Gokaslan, Aaron and Wijmans, Erik and Maksymets, Oleksandr and Clegg, Alex and Turner, John and Undersander, Eric and Galuba, Wojciech and Westbury, Andrew and Chang, Angel X and others},
  journal={arXiv preprint arXiv:2109.08238},
  year={2021}
}

@inproceedings{chang2017matterport3d,
  title={Matterport3D: Learning from RGB-D Data in Indoor Environments},
  author={Chang, Angel and Dai, Angela and Funkhouser, Thomas and Halber, Maciej and Niessner, Matthias and Savva, Manolis and Song, Shuran and Zeng, Andy and Zhang, Yinda},
  booktitle={International Conference on 3D Vision (3DV)},
  year={2017}
}

@article{chiang2024mobility,
  title={Mobility vla: Multimodal instruction navigation with long-context vlms and topological graphs},
  author={Chiang, Hao-Tien Lewis and Xu, Zhuo and Fu, Zipeng and Jacob, Mithun George and Zhang, Tingnan and Lee, Tsang-Wei Edward and Yu, Wenhao and Schenck, Connor and Rendleman, David and Shah, Dhruv and others},
  journal={arXiv preprint arXiv:2407.07775},
  year={2024}
}

@article{chen2025astra,
  title={Astra: Toward General-Purpose Mobile Robots via Hierarchical Multimodal Learning},
  author={Chen, Sheng and He, Peiyu and Hu, Jiaxin and Liu, Ziyang and Wang, Yansheng and Xu, Tao and Zhang, Chi and Zhang, Chongchong and An, Chao and Cai, Shiyu and others},
  journal={arXiv preprint arXiv:2506.06205},
  year={2025}
}

@article{zhou2025fsr,
  title={FSR-VLN: Fast and Slow Reasoning for Vision-Language Navigation with Hierarchical Multi-modal Scene Graph},
  author={Zhou, Xiaolin and Xiao, Tingyang and Liu, Liu and Wang, Yucheng and Chen, Maiyue and Meng, Xinrui and Wang, Xinjie and Feng, Wei and Sui, Wei and Su, Zhizhong},
  journal={arXiv preprint arXiv:2509.13733},
  year={2025}
}

@inproceedings{gu2024conceptgraphs,
  title={Conceptgraphs: Open-vocabulary 3d scene graphs for perception and planning},
  author={Gu, Qiao and Kuwajerwala, Ali and Morin, Sacha and Jatavallabhula, Krishna Murthy and Sen, Bipasha and Agarwal, Aditya and Rivera, Corban and Paul, William and Ellis, Kirsty and Chellappa, Rama and others},
  booktitle={2024 IEEE International Conference on Robotics and Automation (ICRA)},
  pages={5021--5028},
  year={2024},
  organization={IEEE}
}

@inproceedings{cao2021tare,
  title     = {{TARE}: A Hierarchical Framework for Efficiently Exploring Complex {3D} Environments},
  author    = {Cao, Chao and Zhu, Hongbiao and Choset, Howie and Zhang, Ji},
  booktitle = {Proceedings of Robotics: Science and Systems (RSS)},
  year      = {2021},
  month     = {July},
  doi       = {10.15607/RSS.2021.XVII.002}
}

@article{abot-n0,
  title={ABot-N0: Technical Report on the VLA Foundation Model for Versatile Embodied Navigation},
  author={Hu, Junjun and Chen, Jintao and Luo, Minghua and others},
  journal={arXiv preprint arXiv:2602.11598},
  year={2026}
}

@article{astranav-world,
  title={AstraNav-World: World Model for Foresight Control and Consistency},
  author={Chen, Jintao and Hu, Junjun and Chen, Ziyi and Liu, Fei and Chu, Zedong and Wu, Xiaolong and Xu, Mu and Zhang, Shanghang},
  journal={arXiv preprint arXiv:2512.21714},
  year={2025}
}

@article{nav-r2,
  title={Nav-$R^2$ Dual-Relation Reasoning for Generalizable Open-Vocabulary Object-Goal Navigation},
  author={Qiu, Weiming and Cheng, Ze and Wang, Yifan and Xu, Tonghua and Lu, Shijia and Liu, Junhao and Tan, Ping and Qin, Zhengyu},
  journal={arXiv preprint arXiv:2512.02400},
  year={2025}
}

@article{navforesee,
  title={NavForesee: A Unified Vision-Language World Model for Hierarchical Planning and Dual-Horizon Navigation Prediction},
  author={Chen, Jintao and Hu, Junjun and Zhang, Shanghang},
  journal={arXiv preprint arXiv:2512.01550},
  year={2025}
}

@article{socialnav,
  title={SocialNav: Training Human-Inspired Foundation Model for Socially-Aware Embodied Navigation},
  author={Xue, Xinda and Hu, Junjun and Luo, Minghua and Wu, Sichao and Chen, Jintao and others},
  journal={arXiv preprint arXiv:2511.21135},
  year={2025}
}

@article{omninav,
  title={OmniNav: A Unified Framework for Prospective Exploration and Visual-Language Navigation},
  author={Xue, Xinda and Hu, Junjun and Luo, Minghua and Wu, Sichao and others},
  journal={arXiv preprint arXiv:2509.25687},
  year={2025}
}

@article{ce-nav,
  title={CE-Nav: Flow-Guided Reinforcement Refinement for Cross-Embodiment Local Navigation},
  author={Yang, Kai and Li, Tianlin and Xiao, Haoran and Wang, Haonan and others},
  journal={arXiv preprint arXiv:2509.23203},
  year={2025}
}

@article{wang2025cogniplan,
  title={Cogniplan: Uncertainty-guided path planning with conditional generative layout prediction},
  author={Wang, Yizhuo and He, Haodong and Liang, Jingsong and Cao, Yuhong and Chakraborty, Ritabrata and Sartoretti, Guillaume},
  journal={arXiv preprint arXiv:2508.03027},
  year={2025}
}

@inproceedings{shi2025smartway,
  title={Smartway: Enhanced waypoint prediction and backtracking for zero-shot vision-and-language navigation},
  author={Shi, Xiangyu and Li, Zerui and Lyu, Wenqi and Xia, Jiatong and Dayoub, Feras and Qiao, Yanyuan and Wu, Qi},
  booktitle={2025 IEEE/RSJ International Conference on Intelligent Robots and Systems (IROS)},
  pages={16923--16930},
  year={2025},
  organization={IEEE}
}

@article{li2025gvd,
  title={GVD-TG: Topological Graph based on Fast Hierarchical GVD Sampling for Robot Exploration},
  author={Li, Yanbin and Xiao, Canran and Yuan, Shenghai and Yu, Peilai and Li, Ziruo and Zhang, Zhiguo and Chi, Wenzheng and Zhang, Wei},
  journal={arXiv preprint arXiv:2511.18708},
  year={2025}
}

@inproceedings{baek2025pipe,
  title={Pipe planner: Pathwise information gain with map predictions for indoor robot exploration},
  author={Baek, Seungjae and Moon, Brady and Kim, Seungchan and Cao, Muqing and Ho, Cherie and Scherer, Sebastian and Jeon, Jeong Hwan},
  booktitle={2025 IEEE/RSJ International Conference on Intelligent Robots and Systems (IROS)},
  pages={7684--7691},
  year={2025},
  organization={IEEE}
}

@inproceedings{niu2025skeleton,
  title={A skeleton-based topological planner for exploration in complex unknown environments},
  author={Niu, Haochen and Ji, Xingwu and Zhang, Lantao and Wen, Fei and Ying, Rendong and Liu, Peilin},
  booktitle={2025 IEEE International Conference on Robotics and Automation (ICRA)},
  pages={11766--11772},
  year={2025},
  organization={IEEE}
}

@inproceedings{song2025p,
  title={P 2 Explore: Efficient Exploration in Unknown Cluttered Environment with Floor Plan Prediction},
  author={Song, Kun and Chen, Gaoming and Tomizuka, Masayoshi and Zhan, Wei and Xiong, Zhenhua and Ding, Mingyu},
  booktitle={2025 IEEE/RSJ International Conference on Intelligent Robots and Systems (IROS)},
  pages={13090--13096},
  year={2025},
  organization={IEEE}
}

@inproceedings{alama2025rayfronts,
  title={RayFronts: Open-set semantic ray frontiers for online scene understanding and exploration},
  author={Alama, Omar and Bhattacharya, Avigyan and He, Haoyang and Kim, Seungchan and Qiu, Yuheng and Wang, Wenshan and Ho, Cherie and Keetha, Nikhil and Scherer, Sebastian},
  booktitle={2025 IEEE/RSJ International Conference on Intelligent Robots and Systems (IROS)},
  pages={5930--5937},
  year={2025},
  organization={IEEE}
}

@inproceedings{hong2022bridging,
  title={Bridging the gap between learning in discrete and continuous environments for vision-and-language navigation},
  author={Hong, Yicong and Wang, Zun and Wu, Qi and Gould, Stephen},
  booktitle={Proceedings of the IEEE/CVF conference on computer vision and pattern recognition},
  pages={15439--15449},
  year={2022}
}

@article{liu2025fine,
  title={Fine-Grained Instruction-Guided Graph Reasoning for Vision-and-Language Navigation},
  author={Liu, Yaohua and Song, Xinyuan and Deng, Yunfu and Xie, Yifan and Ou, Binkai and Zhong, Yan},
  journal={arXiv preprint arXiv:2503.11006},
  year={2025}
}

@article{huang2025msgnav,
  title={MSGNav: Unleashing the Power of Multi-modal 3D Scene Graph for Zero-Shot Embodied Navigation},
  author={Huang, Xun and Zhao, Shijia and Wang, Yunxiang and Lu, Xin and Zhang, Wanfa and Qu, Rongsheng and Li, Weixin and Wang, Yunhong and Wen, Chenglu},
  journal={arXiv preprint arXiv:2511.10376},
  year={2025}
}

@article{yin2024sg,
  title={Sg-nav: Online 3d scene graph prompting for llm-based zero-shot object navigation},
  author={Yin, Hang and Xu, Xiuwei and Wu, Zhenyu and Zhou, Jie and Lu, Jiwen},
  journal={Advances in neural information processing systems},
  volume={37},
  pages={5285--5307},
  year={2024}
}

@article{chu2026abot,
  title={ABot-N0: Technical Report on the VLA Foundation Model for Versatile Embodied Navigation},
  author={Chu, Zedong and Xie, Shichao and Wu, Xiaolong and Shen, Yanfen and Luo, Minghua and Wang, Zhengbo and Liu, Fei and Leng, Xiaoxu and Hu, Junjun and Yin, Mingyang and others},
  journal={arXiv preprint arXiv:2602.11598},
  year={2026}
}

@article{liu2025navforesee,
  title={NavForesee: A Unified Vision-Language World Model for Hierarchical Planning and Dual-Horizon Navigation Prediction},
  author={Liu, Fei and Xie, Shichao and Luo, Minghua and Chu, Zedong and Hu, Junjun and Wu, Xiaolong and Xu, Mu},
  journal={arXiv preprint arXiv:2512.01550},
  year={2025}
}

@article{miao2025towards,
  title={Towards Physically Executable 3D Gaussian for Embodied Navigation},
  author={Miao, Bingchen and Wei, Rong and Ge, Zhiqi and Gao, Shiqi and Zhu, Jingzhe and Wang, Renhan and Tang, Siliang and Xiao, Jun and Tang, Rui and Li, Juncheng and others},
  journal={arXiv preprint arXiv:2510.21307},
  year={2025}
}

\end{document}